\documentclass{article} % For LaTeX2e
\usepackage{iclr2015,times}
\usepackage{hyperref}
\usepackage{url}

\iclrconference
\iclrfinalcopy

\usepackage{natbib}
\usepackage{amsmath}
\usepackage{amsthm}
\usepackage{amssymb}
\usepackage{relsize}

\usepackage{float}
\usepackage[pdftex]{graphicx}
\graphicspath{{figures/}}

\usepackage{algorithm}
\usepackage{algorithmic}

\renewcommand{\vec}[1]{{\mathbf{#1}}}

\newcommand{\DD}{{\cal D}}
\newcommand{\LL}{{\cal L}}

\newcommand{\E}[2]{\mathop{\mathlarger{\mathbb{E}} }_{#1}\left[#2\right]}
\newcommand{\prob}[2]{p\left(#1 \, | \, #2\right)}
\newcommand{\qrob}[2]{q\left(#1 \, | \, #2\right)}

\newcommand{\xVec}{\vec{x}}
\newcommand{\hVec}{\vec{h}}
\newcommand{\yVec}{\vec{y}}
\newcommand{\aVec}{\vec{a}}

\newcommand{\bmx}[0]{\begin{bmatrix}}
\newcommand{\emx}[0]{\end{bmatrix}}

\title{Reweighted Wake-Sleep}

\author{
J\"org Bornschein and 
Yoshua Bengio \thanks{J\"org Bornschein is a CIFAR Global Scholar; Yoshua Bengio is a CIFAR Senior Fellow} \\
Department of Computer Science and Operations Research\\
University of Montreal\\
Montreal, Quebec, Canada\\
}

\begin{document}

\maketitle

\begin{abstract}
Training deep directed graphical models with many hidden variables and performing inference
remains a major challenge. Helmholtz machines and deep belief networks are such
models, and the wake-sleep algorithm has been proposed to train them.
The wake-sleep algorithm relies on training not just the directed generative model
but also a conditional generative model (the {\em inference network})
that runs backward from visible to latent,
estimating the posterior distribution of latent given visible.
We propose a novel interpretation of the wake-sleep algorithm which suggests
that better estimators of the gradient can be obtained by sampling latent
variables multiple times from the inference network.
This view is based on importance sampling as an estimator of the likelihood,
with the approximate inference network as a proposal distribution. 
This interpretation is confirmed experimentally, showing that better likelihood
can be achieved with this {\em reweighted wake-sleep} procedure.
Based on this interpretation, we propose that a sigmoidal belief network is not
sufficiently powerful for the layers of the inference network in order to
recover a good estimator of the posterior distribution of latent variables. Our
experiments show that using a more powerful layer model, such as NADE, yields
substantially better generative models.
\end{abstract}

%%%%%%%%%%%%%%%%%%%%%%%%%%%%%%%%%%%%%%%%%%%%%%%%%%%%%%%%%%%%%%%%%%%%%%%%%%%%%%%%

\section{Introduction}

Training directed graphical models -- especially models with multiple layers
of hidden variables -- remains a major challenge.  This is unfortunate
because, as has been argued previously~\citep{Hinton06,Bengio-2009-book}, a
deeper generative model has the potential to capture high-level
abstractions and thus generalize better. The exact log-likelihood gradient
is intractable, be it for Helmholtz machines~\citep{Hinton95,Dayan-et-al-1995},
sigmoidal belief networks (SBNs), or deep belief networks
(DBNs)~\citep{Hinton06}, which are directed models with a restricted Boltzmann machine (RBM) as top-layer.
Even obtaining an unbiased estimator of the gradient of the DBN
or Helmholtz machine log-likelihood is not something that has been achieved in
the past. Here we show that it is possible to get an unbiased estimator of
the likelihood (which unfortunately makes it a slightly biased estimator of
the log-likelihood), using an importance sampling approach. Past proposals
to train Helmholtz machines and DBNs rely on maximizing a variational bound
as proxy for the log-likelihood~\citep{Hinton95,Kingma+Welling-ICLR2014,Rezende-et-al-ICML2014}.
The first of these is the {\em wake-sleep} algorithm~\citep{Hinton95}, which 
relies on combining a ``recognition'' network (which we call an {\em approximate
inference network}, here, or simply {\em inference network}) with a generative network.
In the wake-sleep algorithm, they basically provide targets for each other.
We review these previous approaches and introduce a novel approach that
generalizes the wake-sleep algorithm. Whereas the original justification
of the wake-sleep algorithm has been questioned (because we are optimizing
a KL-divergence in the wrong direction), a contribution of this paper is
to shed a different light on the wake-sleep algorithm, viewing it as a special case of
the proposed reweighted wake-sleep (RWS) algorithm, i.e., as reweighted
wake-sleep with a single sample. 
This shows that wake-sleep corresponds to optimizing
a somewhat biased estimator of the likelihood
gradient, while using more samples makes the estimator
less biased (and asymptotically unbiased as more samples are considered).
We empirically show that effect, with clearly better results obtained with $K=5$ samples
than with $K=1$ (wake-sleep), and 5 or 10 being sufficient to achieve good results.
Unlike in the case of DBMs, which rely on a Markov chain to get samples
and estimate the gradient by a mean over those samples, here the samples
are i.i.d., avoiding the very serious problem of mixing between modes that
can plague MCMC methods~\citep{Bengio-et-al-ICML2013} when training
undirected graphical models.

Another contribution of this paper regards the architecture of the deep
approximate inference network. We view the inference network as estimating the
posterior distribution of latent variables given the observed input. With this
view it is plausible that the classical architecture of the inference network
(a SBN, details below) is inappropriate and we test this hypothesis
empirically. We find that more powerful parametrizations that can represent 
non-factorial posterior distributions yield better results.
% each layer as a conditional probability model yields better results.
%In the classical sigmoidal belief network (SBN) (e.g., in the DBN
%and Helmholtz machine), the conditional distribution of each layer of the
%inference network, given the previous layer, is a factorized Bernoulli (where
%the probability for each bit is computed as in a logistic regression with the
%previous layer bits as inputs). 
\iffalse
Here we propose a method to train and evaluate directed graphical models
with binary latent variables according to their estimated log likelihood.
In contrast to most previously proposed methods we do not optimize a
variational bound.  We use an importance sampling based estimator to
perform inference and to compute learning gradients.  The proposal
distribution has the form of a multilayer directed network that runs in
parallel but inverted to the generative network $p(x,h)$.  In parallel to
learning the generative model $p(x)$, we train the proposal distribution
$q(\xVec|\hVec)$ to perform inference / provide a minimum variance
estimator $p(x)$.
\fi

\section{Reweighted Wake-Sleep}

\subsection{The Wake-Sleep Algorithm}

The wake-sleep algorithm was proposed as a way to train Helmholtz machines,
which are deep directed graphical models $p(\xVec,\hVec)$ over visible variables $\xVec$
and latent variables $\hVec$, where the latent variables are organized in
layers $\hVec_k$. %, with the $k$-th layer taking as input the random vector generated by the previous layer in the generating sequence, $\hVec_{k+1}$.
In the Helmholtz machine~\citep{Hinton95,Dayan-et-al-1995}, the top layer
$\hVec_L$ has a factorized unconditional distribution, so that ancestral
sampling can proceed from $\hVec_L$ down to $\hVec_1$ and then the generated
sample $\xVec$ is generated by the bottom layer, given $\hVec_1$. In the
deep belief network (DBN)~\citep{Hinton06}, the top layer is instead
generated by a RBM, i.e., by a Markov chain,
while simple ancestral sampling is used for the others.  Each intermediate
layer is specified by a conditional distribution parametrized 
as a stochastic sigmoidal layer (see section~\ref{sec:SBN} for details).

The wake-sleep algorithm is a training procedure for such generative
models, which involves training an auxiliary network, called the
{\em inference network}, that takes a visible vector $\xVec$ as input and
stochastically outputs samples $\hVec_k$ for all layers $k=1$ to $L$. 
The inference network outputs samples from a distribution that
should estimate the conditional probability of the latent variables of the
generative model (at all layers) given the input. Note that in these kinds of directed models 
%(and this is generally the case with latent variables), 
exact inference, i.e., sampling from $p(\hVec|\xVec)$ is intractable.

The wake-sleep algorithm proceeds in two phases. In the {\em wake phase},
an observation $\xVec$ is sampled from the training distribution $\DD$ and
propagated stochastically up the inference network (one layer at a time), thus sampling latent
values $\hVec$ from $q(\hVec | \xVec)$. Together with $\xVec$, the sampled $\hVec$
forms a target for training $p$, i.e., one performs a step of gradient ascent update with respect to maximum
likelihood over the generative model $p(\xVec, \hVec)$, with the data $\xVec$ and the
inferred $\hVec$. This is useful
because whereas computing the gradient of the marginal likelihood
$p(\xVec)=\sum_\hVec p(\xVec,\hVec)$ is intractable, 
computing the gradient of the complete log-likelihood $\log p(\xVec,\hVec)$
is easy. In addition, these updates decouple all the layers (because both the input and the target of each layer
are considered observed). In the {\em sleep-phase}, a ``dream'' sample is obtained from the
generative network by ancestral sampling from $p(\xVec, \hVec)$ and is used
as a target for the maximum likelihood training of the inference
network, i.e., $q$ is trained to estimate $p(\hVec | \xVec)$.

The justification for the wake-sleep algorithm that was originally proposed is based on the following variational bound,
\[
  \log p(\xVec) \geq \sum_{\hVec} q(\hVec|\xVec) \log \frac{p(\xVec,\hVec)}{q(\hVec|\xVec)}
\]
that is true for any inference network $q$, but the bound becomes tight as $q(\hVec|\xVec)$ approaches $p(\hVec|\xVec)$.
Maximizing this bound with respect to $p$ corresponds to the wake phase update.
The update with respect to $q$ should minimize $KL(q(\hVec|\xVec) || p(\hVec|\xVec))$ (with $q$ as the reference)
but instead the sleep phase update minimizes the reversed KL divergence, $KL(p(\hVec|\xVec) || q(\hVec|\xVec))$
(with $p$ as the reference).

\subsection{An Importance Sampling View yields Reweighted Wake-Sleep}
\label{sec:pxest}

If we think of $q(\hVec|\xVec)$ as estimating $p(\hVec|\xVec)$ and train it accordingly
(which is basically what the sleep phase of wake-sleep does), then we can reformulate
the likelihood as an importance-weighted average:
%
%\begin{align}  
%  p(\xVec) &= \sum_{\hVec} p(\xVec, \hVec) = \sum_{\hVec} \qrob{\hVec}{\xVec}
%                   \frac{p(\xVec, \hVec)}{\qrob{\hVec}{\xVec}} \nonumber \\
%                &= \E{\hVec \sim \qrob{\hVec}{\xVec}}{\frac{p(\xVec, \hVec)}{\qrob{\hVec}{\xVec}}} 
%                \simeq \frac{1}{K} \sum_{\substack{k=1 \\ \hVec^{(k)} \sim \qrob{\hVec}{\xVec}} }^K
%                   \frac{p(\xVec, \hVec^{(k)})}{\qrob{\hVec^{(k)}}{\xVec}}          \label{eq:pxest}
%\end{align}
\begin{align}  
  p(\xVec) %&= \sum_{\hVec} p(\xVec, \hVec) 
            = \sum_{\hVec} \qrob{\hVec}{\xVec} \frac{p(\xVec, \hVec)}{\qrob{\hVec}{\xVec}}
                &= \E{\hVec \sim \qrob{\hVec}{\xVec}}{\frac{p(\xVec, \hVec)}{\qrob{\hVec}{\xVec}}} 
                \simeq \frac{1}{K} \hspace{-0.3cm} \sum_{\substack{k=1 \\ \hVec^{(k)} \sim \qrob{\hVec}{\xVec}} }^K \hspace{-0.3cm} 
                   \frac{p(\xVec, \hVec^{(k)})}{\qrob{\hVec^{(k)}}{\xVec}}          \label{eq:pxest}
\end{align}
Eqn. \eqref{eq:pxest} is a consistent and unbiased estimator for the marginal
likelihood $p(\xVec)$.  The optimal $q$ that results in a minimum variance
estimator is $q^{*}(\hVec\,|\,\xVec)=\prob{\hVec}{\xVec}$.  In fact we can show
that this is a zero-variance estimator, i.e., the best possible one that will
result in a perfect $p(x)$ estimate even with a single arbitrary sample 
$\hVec \sim \prob{\hVec}{\xVec}$: 
\begin{align}
 \E{h \sim p(\hVec\,|\,\xVec)}
      {\frac{\prob{\hVec}{\xVec}p(\xVec)}{p(\hVec\,|\,\xVec)}} 
%      {\frac{p(\xVec, \hVec)}{p(\hVec\,|\,\xVec)}} 
    = p(\xVec) \; \E{h \sim p(\hVec\,|\,\xVec)}{1} = p(\xVec)
\end{align}
Any mismatch between $q$ and $\prob{\hVec}{\xVec}$ will increase the variance
of this estimator, but it will not introduce any bias.  In practice however, we
are typically interested in an estimator for the {\em log}-likelihood. Taking
the logarithm of \eqref{eq:pxest} and averaging over multiple datapoints will
result in a conservative biased estimate and will, on average, underestimate
the true log-likelihood due to the concavity of the logarithm. Increasing the
number of samples will decrease both the bias and the variance.  Variants of
this estimator have been used in e.g. \citep{Rezende-et-al-ICML2014,KarolAndriyDaan14}
to evaluate trained models.

\iffalse
for set of datapoints $\DD=\xVec^{(n)}$. By Jensens's
inequality, we get 
In this case, in expectation over samples,
we get a lower bound on the log-likelihood, by Jensen's inequality: the
expected value over samples of the log of the importance weighted likelihood
estimator is less or equal than the log of the expected value, i.e., less or
equal to the true log-likelihood.
 This is good because it gives us a conservative
estimator of the log-likelihood, in average, i.e., it tends to underestimate the
ground truth.
From the point of view of the log-likelihood gradient, it means
that our estimator is the gradient of a lower bound on the log-likelihood.
This is also true of variational methods such as described below (section~\ref{sec:var-AE}).
However, unlike with these methods, here {\em the bound can be made 
arbitrarily tighter by simply using more samples}, because the inner
average over samples (before applying the log) converges to its expectation.

\begin{align}
  \label{eq:llbound}
  \text{LL bound} &=
  avg_{\xVec \in \DD} \; \E{\hVec^{(k)}\sim q(\hVec|\xVec)}
    {\log p(\xVec, \hVec) - \log q(\hVec | \xVec)} \\
  \label{eq:llest}
  \text{LL} &=
  avg_{\xVec \in \DD} \log \left( \E{\hVec^{(k)}\sim q(\hVec|\xVec)}
    {\frac{ p(\xVec, \hVec) }{ q(\hVec | \xVec)}} \right)  \\
  \text{LL bound} &\leq   \text{LL}
\end{align}

\fi

%%%%%%%%%%%%%%%%%%%%%%%%%%%%%%%%%%%%%%%%%%%%%%%%%%%%%%%%%%%%%%%%%%%%%%%%%%%%%
%
\subsection{Training by Reweighted Wake-Sleep}

We now consider the models $p$ and $q$ parameterized with parameters $\theta$ and $\phi$ respectively.

%\subsubsection{Updating $p_{\theta}$ for given $q_\phi$}
{\bf Updating $p_{\theta}$ for given $q_\phi$}:
We propose to use an importance sampling estimator based on eq. \eqref{eq:pxest}
to compute the gradient of the marginal log-likelihood 
$\LL_p(\theta, \xVec) = \log p_{\theta}(\xVec)$: 
\begin{align}
  \frac{\partial}{\partial \theta} \LL_p(\theta, \xVec \sim \DD) 
%    &= \frac{\partial}{\partial \theta} \log p_{\theta}(\xVec) 
%     = \frac{1}{p(\xVec)} \frac{\partial}{\partial \theta} p(\xVec) \\
%     = \frac{1}{p(\xVec)} \frac{\partial}{\partial \theta} \sum_{\hVec} p(\xVec, \hVec) \nonumber \\
%    &= \frac{1}{p(\xVec)} \sum_{\hVec} p(\xVec, \hVec) 
%            \frac{\partial}{\partial \theta} \log  p(\xVec, \hVec) \nonumber \\
%    &= \frac{1}{p(\xVec)} \sum_{\hVec} \qrob{\hVec}{\xVec} \frac{p(\xVec, \hVec)}{\qrob{\hVec}{\xVec}}
%            \frac{\partial}{\partial \theta} \log  p(\xVec, \hVec) \nonumber \\
    &= \frac{1}{p(\xVec)} \E{\hVec \sim \qrob{\hVec}{\xVec}}{\frac{p(\xVec, \hVec)}{\qrob{\hVec}{\xVec}}
            \frac{\partial}{\partial \theta} \log p(\xVec, \hVec)} \nonumber \\
  \label{eq:grad_p}
%    &\simeq \frac{1}{\sum_k \omega_k} \sum_{k=1}^K \omega_k
%            \frac{\partial}{\partial \theta} \log p(\xVec, \hVec^{(k)}) \\
    &\simeq \sum_{k=1}^K \tilde{\omega}_k \frac{\partial}{\partial \theta} \log p(\xVec, \hVec^{(k)}) 
    \;\;\text{with}\; \hVec^{(k)} \sim \qrob{\hVec}{\xVec}  \\ 
 \text{and the importance} & \text{ weights}\; 
    \tilde{\omega}_k = \frac{\omega_k}{\sum_{k'=1}^K \omega_{k'}} ; \;  
    \omega_k = \frac{p(\xVec, \hVec^{(k)})}{\qrob{\hVec^{(k)}}{\xVec}} \text{.}\nonumber
    %&= \frac{1}{p(\xVec)} \sum_{\hVec} \prob{\hVec}{\xVec}
\end{align}
See the supplement for a detailed derivation.
Note that this is a biased estimator because it implicitly contains a division by the estimated $p(x)$.
Furthermore, there is no guarantee that $q=\prob{\hVec}{\xVec}$ results in a minimum
variance estimate of this gradient. But both, the bias and the variance, decrease as the number of samples 
is increased. Also note that the wake-sleep algorithm uses a gradient that is equivalent to using only $K=1$ sample.
Another noteworthy detail about eq. \eqref{eq:grad_p} is that the importance weights $\tilde{\omega}$ are automatically 
normalized such that they sum up to one.

%Equation \eqref{eq:grad_p} is a biased but consistent (asymptotically unbiased)
%.estimator of the gradient.

{\bf Updating $q_{\phi}$ for given $p_\theta$}:
In order to minimize the variance of the estimator \eqref{eq:pxest} we would like
$\qrob{\hVec}{\xVec}$ to track $\prob{\hVec}{\xVec}$. We propose to train $q$ using maximum 
likelihood learning with the loss $\LL_q(\phi, \xVec, \hVec) = \log q_\phi(\xVec|\hVec)$.
There are at least two reasonable options how to obtain training data for $\LL_q$: % $(\xVec, \hVec)$ pairs:
1) maximize $\LL_q$ under the empirical training distribution $\xVec \sim {\cal D}$, $\hVec \sim \prob{\hVec}{\xVec}$, or
2) maximize $\LL_q$ under the generative model $(\xVec, \hVec) \sim p_{\theta}(\xVec, \hVec)$.
We will refer to the former as {\em wake phase q-update} and to the latter as {\em sleep phase q-update}.
In the case of a DBN (where the top layer is generated by an RBM), there is an intermediate solution called 
contrastive-wake-sleep, which has been proposed in~\citep{Hinton06}. 
In contrastive wake-sleep we sample $\xVec$ from the training distribution, propagate it stochastically
into the top layer and use that $\hVec$ as starting point for a short Markov chain in the RBM, then 
sample the other layers in the generative network $p$ to generate the rest of $(\xVec,\hVec)$. The
objective is to put the inference network's capacity where it matters most, i.e., near the input configurations
that are seen in the training set. 

Analogous to eqn. \eqref{eq:pxest} and \eqref{eq:grad_p} we use importance sampling
to derive gradients for the wake phase q-update: 
\begin{align}
  \frac{\partial}{\partial \phi} \LL_q(\phi, \xVec \sim \DD) 
%    &= \frac{\partial}{\partial \phi} \sum_{\hVec} 
%        p(\xVec, \hVec) {\log q_{\phi}(\hVec | \xVec)}  \nonumber \\
%    &= \frac{1}{p(\xVec)} \E{\hVec \sim \qrob{\hVec}{\xVec}}{\frac{p(\xVec, \hVec)}{\qrob{\hVec}{\xVec}}
%            \frac{\partial}{\partial \phi} \log q_{\phi}(\hVec | \xVec)}  \nonumber \\
    \label{eq:grad_q}
%    &\simeq \frac{1}{\sum_k \omega_k} \sum_{k=1}^K \omega_k
%            \frac{\partial}{\partial \phi} \log q_{\phi}(\hVec^{(k)} | \xVec)
    &\simeq \sum_{k=1}^K \tilde{\omega}_k \frac{\partial}{\partial \phi} \log q_{\phi}(\hVec^{(k)} | \xVec)
%\text{with}\;&\omega_k = \frac{p(\xVec, \hVec^{(k)})}{\qrob{\hVec^{(k)}}{\xVec}} \;
%\text{and}\; \hVec^{(k)} \sim \qrob{\hVec}{\xVec} \nonumber
%    &= \cdots \\
%    &= \E{\xVec \sim \DD}{ \E{\hVec \sim \qrob{\hVec}{\xVec)}} 
%     { \omega \; \frac{\partial}{\partial \phi} \log \qrob{\hVec}{\xVec}} }
%     \; \text{with the importance weights } \omega = \frac{\prob{\hVec}{\xVec}}{\qrob{\hVec}{\xVec}}
\end{align}
with the same importance weights $\tilde{\omega}_k$ as in \eqref{eq:grad_p} 
(the details of the derivation can again be found in the supplement).
Note that this is equivalent to optimizing $q$ so as to minimize $KL(p(\cdot|\xVec)
\parallel q(\cdot|\xVec))$. 
For the sleep phase q-update we consider the model distribution $p(\xVec, \hVec)$ a fully observed system 
and can thus derive gradients without further sampling:
\begin{align}
  \frac{\partial}{\partial \phi} \LL_q(\phi, (\xVec, \hVec) ) 
%  &= \frac{\partial}{\partial \phi} {\log q_{\phi}(\hVec | \xVec)}  \\
%  &= \E{\xVec, \hVec \sim p(\xVec, \hVec)}{
%     \frac{\partial}{\partial \phi} {\log \qrob{\hVec}{\xVec} } } \\
%  &\simeq 
    \label{eq:grad_q_sleep}
  & = \frac{\partial}{\partial \phi} \log q_{\phi}(\hVec | \xVec) 
    \text{ with } \xVec, \hVec \sim p(\xVec, \hVec)
\end{align}
This update is equivalent to the sleep phase update in the classical wake-sleep algorithm.

\begin{algorithm}[ht]
\caption{Reweighted Wake-Sleep training procedure and likelihood estimator.
$K$ is the number of approximate inference samples and controls the trade-off
between computation and accuracy of the estimators (both for the gradient
and for the likelihood). We typically use a large value ($K=100,000$) for test set likelihood
estimator but a small value ($K=5$) for estimating gradients. Both the
wake phase and sleep phase update rules for $q$ are optionally included (either one or both can be used,
and best results were obtained using both).
The original wake-sleep algorithm has $K$=1 and only uses the sleep phase update of $q$. 
To estimate the log-likelihood at test time, only the computations up to $\widehat{\LL}$ are required.
% to estimate the log-likelihood, and a larger value of $K$ (500 in the experiments) is preferable to obtain a more accurate estimator.
}
\begin{algorithmic}
\label{alg:RWS}
\FOR{number of training iterations}
  \STATE{$\bullet$ Sample example(s) $\xVec$ from the training distribution}
  \FOR{$k=1$ to $K$}
    \STATE{$\bullet$ Layerwise sample latent variables $\hVec^{(k)}$ from $q(\hVec|\xVec)$}
     %(first layer above $\xVec$, second layer, etc. up to top hidden layer).}
    \STATE{$\bullet$ Compute $q(\hVec^{(k)}|\xVec)$ and $p(\xVec,\hVec^{(k)})$}
  \ENDFOR
  \STATE{$\bullet$ Compute unnormalized weights $\omega_k = \frac{p(\xVec, \hVec^{(k)})}{\qrob{\hVec^{(k)}}{\xVec}}$}
  \STATE{$\bullet$ Normalize the weights $\tilde{\omega}_k = \frac{\omega_k}{\sum_{k'} \omega_{k'}}$}
  \STATE{$\bullet$ Compute unbiased likelihood estimator $\hat{p}(\xVec) = {\rm average}_k \; \omega_k$}
  \STATE{$\bullet$ Compute log-likelihood estimator $\widehat{\LL}(\xVec) = \log {\rm average}_k \; \omega_k$}
  %\STATE{$\bullet$ {\bf Wake-phase update of $p$}. Compute asymptotically unbiased estimator of log-likelihood gradient w.r.t. $p$, 
  %$\sum_k \tilde{\omega}_k \frac{\partial \log p(\xVec,\hVec^{(k)})}{\partial \theta}$,
  %and perform an update of $p$'s parameters using it}
  \STATE{$\bullet$ {\bf Wake-phase update of $p$}: Use gradient estimator $\sum_k \tilde{\omega}_k \frac{\partial \log p(\xVec,\hVec^{(k)})}{\partial \theta}$}
  \STATE{$\bullet$ Optionally, {\bf wake phase update of $q$}: Use gradient estimator $\sum_k \tilde{\omega}_k \frac{\partial \log q(\hVec^{(k)}|\xVec)}{\partial \phi}$}
  \STATE{$\bullet$ Optionally, {\bf sleep phase update of $q$}: Sample $(\xVec',\hVec')$ from $p$ and use gradient $\frac{\partial \log q(\hVec'|\xVec')}{\partial \phi}$}
\ENDFOR
\end{algorithmic}
\end{algorithm}

\subsection{Relation to Wake-Sleep and Variational Bayes}
\label{sec:var-AE}

Recently, there has been a resurgence of interest in algorithms related to the Helmholtz machine 
and to the wake-sleep algorithm for directed graphical models containing either continuous or discrete
latent variables:

In Neural Variational Inference and Learning \citep[NVIL, ][]{Andriy-Karol-2014} the
authors propose to maximize the variational lower bound on the
log-likelihood to get a joint objective for both $p$ and $q$. It
was known that this approach results in a gradient estimate of very high variance
for the recognition network $q$~\citep{Dayan1996varieties}. In the NVIL paper, the authors
therefore propose variance reduction techniques such as {\em baselines}
to obtain a practical algorithm that enhances significantly over the original
wake-sleep algorithm. 
In respect to the computational complexity we note that while we draw $K$ samples from 
the inference network for RWS, NVIL on the other hand draws only a single sample from $q$ but 
maintains, queries and trains an additional auxiliary baseline estimating network. With RWS and a typical value of $K=5$ we thus require at least 
twice as many arithmetic operations, but we do not have to store the baseline network and 
do not have to find suitable hyperparameters for it.

%Compared to NVIL, RWS requires $K$-times as much computations the computations to 
%calculate gradients for $p$ and $q$, but we do not maintain and train an auxiliary
%baseline estimating network. For $K=5$ we thus expect to require roughly twice the 
%computational complexity than NVIL.
% XXX

Recent examples for continuous latent variables include
the auto-encoding variational Bayes~\citep{Kingma+Welling-ICLR2014} and
stochastic backpropagation papers~\citep{Rezende-et-al-ICML2014}. In both
cases one maximizes a variational lower bound on the log-likelihood that is
rewritten as two terms: one that is log-likelihood reconstruction
error through a stochastic encoder (approximate inference) - decoder
(generative model) pair, and one that regularizes the output of the
approximate inference stochastic encoder so that its marginal distribution
matches the generative prior on the latent variables (and the latter is
also trained, to match the marginal of the encoder output).
%
%One of the contributions of this paper is that we find wake-sleep to be a
%special case of reweighted wake-sleep, with a single sample, $K$=1. This makes
%the regular wake-sleep update a clearly biased estimator of the likelihood
%gradient. It also suggests that reweighted wake-sleep can only improve with
%respect to wake-sleep, since both bias and variance must decrease as we
%consider more samples. Experiments described below confirm that hypothesis.
%
Besides the fact that these variational auto-encoders are only for continuous
latent variables, another difference with the reweighted wake-sleep
algorithm proposed here is that in the former, a single sample from the approximate inference
distribution is sufficient to get an unbiased estimator of the gradient
of a proxy (the variational bound). Instead, with the reweighted wake-sleep,
a single sample would correspond to regular wake-sleep, which gives a biased
estimator of the likelihood gradient. On the other hand, as the number of
samples increases, reweighted wake-sleep provides a less biased (asymptotically
unbiased) estimator of the log-likelihood and of its gradient.
Similar in spirit, but aimed at a structured output prediction task is the
method proposed by \cite{Tang+Salakhutdinov-NIPS2013}. The authors optimize the
variational bound of the log-likelihood instead of the direct IS estimate but
they also derive update equations for the proposal distribution that resembles
many of the properties also found in reweighted wake-sleep.

\section{Component Layers}

Although the framework can be readily applied to continuous variables, we
here restrict ourselves to distributions over binary visible and binary
latent variables.  We build our models by combining probabilistic components,
each one associated with one of the layers of the generative network or of the
inference network.  The generative model can therefore be written as 
$p_{\theta}(\xVec, \hVec) = p_0(\xVec|\,\hVec_1) \, p_{1}(\hVec_{1}|\,\hVec_2) \, \cdots \, p_L(\hVec_L)$, 
while the inference network has the form 
$q_\phi(\hVec\,|\,\xVec) = q_1(\hVec_1\,|\,\xVec) \cdots q_L(\hVec_L\,|\,\hVec_{L-1})$.
For a distribution $P$ to be a suitable component we must have a method to efficiently 
compute $P(\xVec^{(k)}|\,\yVec^{(k)})$ given $(\xVec^{(k)}$ , $\yVec^{(k)})$, 
and we must have a method to efficiently draw i.i.d. samples 
$\xVec^{(k)} \sim P(\xVec\,|\,\yVec)$ for a given $\yVec$. 
In the following we will describe experiments containing three kinds of layers:

\label{sec:SBN}
{\bf Sigmoidal Belief Network (SBN) layer:} A SBN layer~\citep{Saul+96} is a directed 
graphical model with independent variables $x_i$ given the parents $\yVec$:
\begin{align}
  P^{\text{SBN}}(x_i=1\,|\,\yVec) = \sigma( W^{i,:} \, \yVec + b_i )
\end{align}
Although a SBN is a very simple generative model given $\yVec$, 
performing inference for $\yVec$ given $\xVec$ is in general intractable.

{\bf Autoregressive SBN layer (AR-SBN, DARN):}
If we consider $x_i$ an ordered set of observed variables and introduce 
directed, autoregressive links between all previous $\xVec_{<i}$ and a given $x_i$, 
we obtain a fully-visible sigmoid belief network 
\cite[FVSBN, ][]{Frey98,Bengio+Bengio-NIPS99}. When we additionally condition a FVSBN 
on the parent layer's $\yVec$ we obtain a layer model that was first used
in Deep AutoRegressive Networks \citep[DARN, ][]{KarolAndriyDaan14}:
%is similar to a SBN layer but with the output units $x_i$
%not being independent of each other, given the layer's input $\yVec$. Instead, their dependency
%is captured by a fully connected directed acyclic graph where the $x_i$
%
\begin{align}
  P^{\text{AR-SBN}}(x_i=1\,|\,\xVec_{<i}, \yVec) = \sigma( W^{i,:} \, \yVec + S^{i,<i} \xVec_{<i} + b_i)
\end{align}
We use $\xVec_{<i} = (x_1, x_2, \cdots, x_{i-1})$ to refer to the vector
containing the first $i$-1 observed variables. The matrix $S$ is a lower
triangular matrix that contains the autoregressive weights between the 
variables $x_i$, and with $S^{i,<j}$ we refer to the first $j$-1 elements of the $i$-th row
of this matrix. 
In contrast to a regular SBN layer, the units $x_i$ are thus not independent of each other but 
can be predicted like in a logistic regression in terms of its predecessors
$\xVec_{<i}$ and of the input of the layer, $\yVec$.

{\bf Conditional NADE layer:}
The Neural Autoregressive Distribution Estimator \citep[NADE, ][]{Larochelle+Murray-2011} is a model
that uses an internal, accumulating hidden layer to predict variables $x_i$ given the vector containing all previously variables $\xVec_{<i}$.
Instead of logistic regression in a FVSBN or an AR-SBN, the dependency between the variables $x_i$ 
is here mediated by an MLP~\citep{Bengio+Bengio-NIPS99}:
%DARN is thus a special case of NADE without a (deterministic) hidden layer to mediate the
%conditional dependency between $x_i$ and its predecessors. Instead of a logistic
%regression, that dependency is mediated 
%
\begin{align}
  P(x_i=1\,|\,\xVec_{<i}) = \sigma( V^{i,:} \sigma( W^{:,<i} \, \xVec_{<i} + \aVec ) + b_i))
\end{align}
With $W$ and $V$ denoting the encoding and decoding matrices for the NADE hidden layer.
For our purposes we condition this model on the random variables $\yVec$:
\begin{align}
  P^{\text{NADE}}(x_i=1\,|\,\xVec_{<i}, \yVec) 
    = \sigma( V^{i,:} \sigma( W^{:,<i} \, \xVec_{<i} + U_a \, \yVec + \aVec ) + U_b^{i,:} \, \yVec + b_i))
\end{align}
Such a conditional NADE has been used previously for modeling musical
sequences~\citep{Boulanger+al-ICML2012-small}.

For each layer distribution we can construct an unconditioned distribution by
removing the conditioning variable $\yVec$. We use such
unconditioned distributions as top layer $p(\hVec)$ for the generative network $p$.
%By removing $\yVec$ from a SBN layer we obtain a factorized Bernoulli
%distribution, by removing $\yVec$ from a AR-SBN layer we obtain a fully-visible 
%sigmoid belief network, and by removing $\yVec$ from a conditional NADE layer we obtain a regular, unconditioned NADE. 

%\subsubsection{parameterizations}

%Given the formula (\ref{learn_p}) and (\ref{learn_q}), there are plenty of options 
%in parameterizing $p_{\theta}(h^i_1,h^i_2,x)$ and $q_{\phi}(h^i_1,h^i_2|x)$. 
%
%For $p_{\theta}$, Sigmoid Belief Networks \citep{Saul+96} is most straightforward. 
%Apart from using the simple Sigmoid Belief Networks,  
%The other choices for those conditionals are conditional NADEs. For the prior $p(h_2)$ 
%there are also many possibilities, such as NADEs, mixture models. 

%For $q_{\phi}$, if we assume it further factorizes as 
%$q_{\phi}(x,h_1,h_2)=q_{\alpha}(h_1|x)q_{\beta}(h_2|h_1)$, then all sorts of 
%parameterizations, such as, conditional NADE, are possible.

%%%%%%%%%%%%%%%%%%%%%%%%%%%%%%%%%%%%%%%%%%%%%%%%%%%%%%%%%%%%%%%%%%%%%%%%%%%%%%%%%%
\section{Experiments}

Here we present a series of experiments on the MNIST and the CalTech-Silhouettes
datasets. The supplement describes additional experiments on 
smaller datasets from the UCI repository. With these experiments we 
1) quantitatively analyze the influence of the number of samples $K$, 2)
demonstrate that using a more powerful layer-model for the inference network
$q$ can significantly enhance the results even when the generative model is a
factorial SBN, and 3) show that we approach state-of-the-art performance when using 
either relatively deep models or when using powerful layer models such as a conditional NADE. 
Our implementation is available at \url{https://github.com/jbornschein/reweighted-ws/}.

\vspace{-0.2cm}
\subsection{MNIST}
\vspace{-0.2cm}

\begin{figure}[t]
  \centering
  \includegraphics[width=\linewidth]{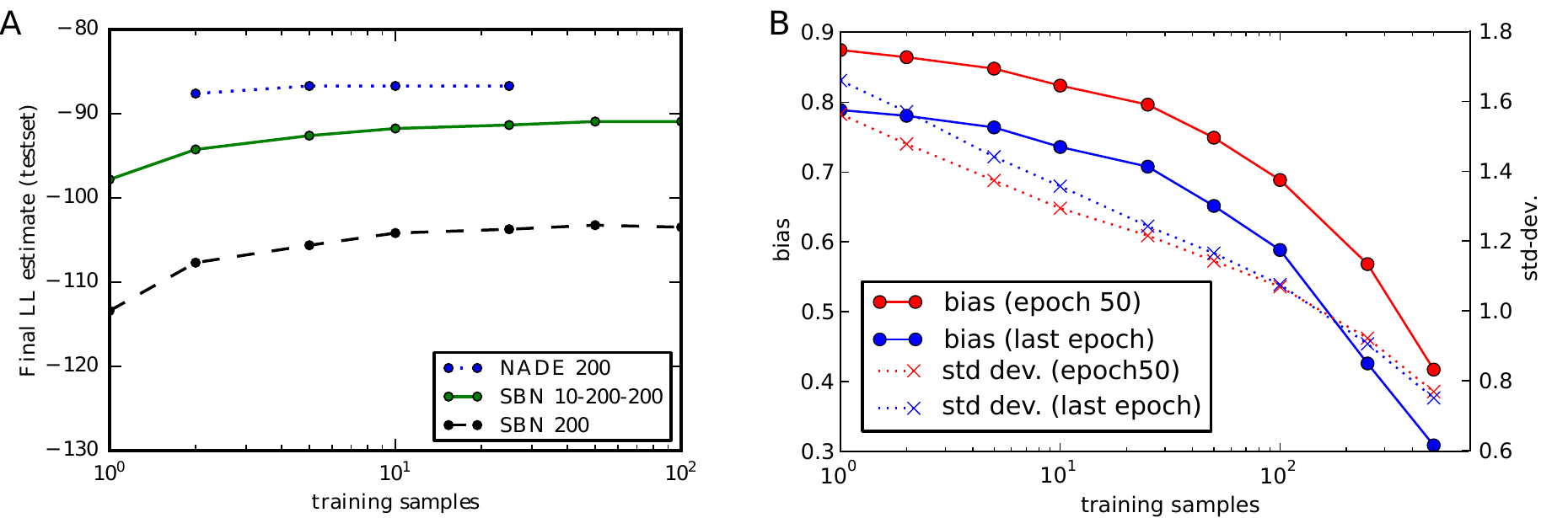}
  \vspace{-0.5cm}
  \caption{
    \label{fig:mnist1}
    {\bf A} Final log-likelihood estimate w.r.t. number of samples used during training.
    {\bf B} $L_2$-norm of the bias and standard deviation of the low-sample estimated 
     $p_\theta$ gradient relative to a high-sample (K=5,000) based estimate.
  }
\end{figure}

\begin{table}
\centering
\small
\begin{tabular}{|l|r||c|c||c|c|}
  \hline 
  \small
          &             & NVIL    & wake-sleep    &          RWS &           RWS \\
  P-model & size        &         &               & Q-model: SBN & Q-model: NADE \\

  \hline 
  SBN    & 200          & (113.1) & 116.3 (120.7) & 103.1 &  95.0 \\
  SBN    & 200-200      &  (99.8) & 106.9 (109.4) &  93.4 &  91.1 \\
  SBN    & 200-200-200  &  (96.7) & 101.3 (104.4) &  90.1 &  88.9 \\
  \hline
  AR-SBN & 200          &       &                 &        &  89.2 \\
  AR-SBN & 200-200      &       &                 &        &  92.8 \\
  \hline
  NADE   & 200          &       &                 &        &  86.8 \\
  NADE   & 200-200      &       &                 &        &  87.6 \\
  \hline 
\end{tabular}
\caption{
MNIST results for various architectures and training methods.  In the 3rd
column we cite the numbers reported by \cite{Andriy-Karol-2014}.  Values in
brackets are variational NLL bounds, values without brackets report NLL estimates
(see section \ref{sec:pxest}). 
}
\label{tab:mnist}
\end{table}

%\begin{table}
%\centering
%\small
%\begin{tabular}{|l|r||c|c||c|c|c|}
%  \hline 
%  \small
%          &      & NVIL         & wake-sleep  & wake-sleep &          RWS &           RWS \\
%  P-model & size & (NLL bound)  & (NLL bound) & (NLL est.) & Q-model: SBN & Q-model: NADE \\
%
%  \hline 
%  SBN    & 200          & 113.1 & 120.7 & 116.3 & 105.7 &  96.7 \\
%  SBN    & 200-200      &  99.8 & 109.4 & 106.9 &  98.3 &  92.3 \\
%  SBN    & 200-200-200  &  96.7 & 104.4 & 101.3 &  94.7 &  91.8 \\
%%  SBN    & 10-200-200   &       &       &  97.8 &  91.9 &  88.9 \\
%  \hline
%  AR-SBN & 200          &       &        &        &        &  91.8 \\
%  AR-SBN & 200-200      &       &        &        &        &  95.2 \\
%  \hline
%  NADE   & 200          &       &        &        &        &  86.8 \\
%  NADE   & 200-200      &       &        &        &        &  87.6 \\
%  \hline 
%\end{tabular}
%\caption{
%MNIST for different architectures and network depths. In the third column 
%we cite the numbers reported by \cite{Andriy-Karol-2014}. Columns three and
%four report the variational NLL bounds; columns 5 to 8 report the NLL 
%estimates (see section \ref{sec:pxest}).
%}
%\label{tab:mnist}
%\end{table}

\begin{figure}[t]
  \centering
  \includegraphics[width=\linewidth]{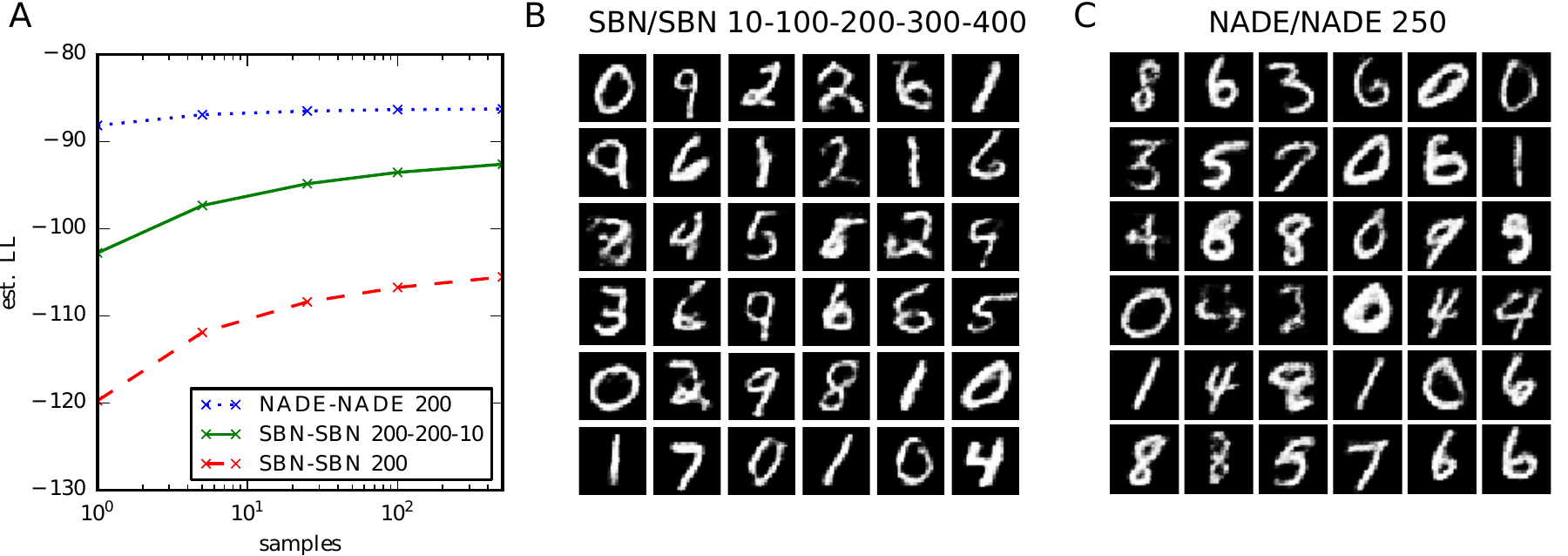}
  \vspace{-0.5cm}
  \caption{
    \label{fig:mnist2}
    {\bf A} Final log-likelihood estimate w.r.t. number of test samples used. % Solid lines: estimated using the importance sampler; dashed-lines: variational {\em bound} estimator.
    {\bf B} Samples from the SBN/SBN 10-200-200 generative model.  
    {\bf C} Samples from the NADE/NADE 250 generative model. 
    (We show the probabilities from which each pixel is sampled)
  }
\end{figure}

\begin{table}[t]
\centering
\small
\begin{tabular}{|l|c|c|}
  \hline 
  \multicolumn{3}{|c|}{\bf Results on binarized MNIST} \\
  \hline 
  \hline 
         & NLL   & NLL \\
  Method & bound & est. \\
  \hline 
  RWS (SBN/SBN 10-100-200-300-400)   &       & 85.48 \\
  RWS (NADE/NADE 250)                &       & 85.23 \\
  RWS (AR-SBN/SBN 500)$\dagger$      &       & 84.18 \\
%  RWS (SBN/NADE 10-200-200)          &       & 88.47 \\
%                                     &       & \footnotesize $\pm$ 0.43 \\
  \hline 
  NADE (500 units, [1])              &       & 88.35 \\
  EoNADE (2hl, 128 orderings, [2])   &       & 85.10 \\
  DARN (500 units, [3])              &       & 84.13 \\
  RBM (500 units, CD3, [4])          & 105.5 & \\
  RBM (500 units, CD25, [4])         & 86.34 & \\
  DBN (500-2000, [5])                & 86.22 & {\footnotesize 84.55} \\
  \hline 
\end{tabular}
\begin{tabular}{|l|c|c|}
  \hline 
  \multicolumn{2}{|c|}{\bf Results on CalTech 101 Silhouettes} \\
  \hline 
  \hline 
                                 & NLL  \\
  Method                         & est. \\
  \hline 
  RWS (SBN/SBN 10-50-100-300)    & 113.3 \\
%  RWS (SBN/NADE 10-50-100-300)   & 112.6 \\
  RWS (NADE/NADE 150)            & 104.3 \\
                                 & \\
  \hline 
  NADE (500 hidden units)        & 110.6 \\
  RBM (4000 hidden units, [6])   & 107.8 \\
                                 &  \\ 
                                 &  \\
                                 &  \\
                                 &  \\
%                                 &  \\
  \hline 
\end{tabular}
\caption{
Various RWS trained models in
relation to previously published methods: [1] \cite{Larochelle+Murray-2011}, [2]
\cite{Uria+al-ICML2014}, [3] \cite{KarolAndriyDaan14}  [4] \cite{SalMurray08}, [5] \cite{MurraySal09}, 
[6] \cite{Cho2013enhanced}. $\dagger$ Same model as the best performing in [3]: a AR-SBN with {\em deterministic}
hidden variables between the observed and latent. All RWS NLL estimates on MNIST have confidence intervals of $\approx \pm 0.40$.
%In row 4 we report 95\% confidence intervals.
%and to state of the art
%(bold). Left: Results on MNIST; Right: results on CalTech 101 Silhouettes. 
}
\label{tab:results}
\end{table}

We use the MNIST dataset that was binarized according to \cite{MurraySal09} and
downloaded in binarized form from~\citep{LarochelleBinarizedMNIST}. For
training we use stochastic gradient decent with momentum ($\beta$=0.95) and set
mini-batch size to 25. The experiments in this paragraph were run with learning
rates of \{0.0003, 0.001, and 0.003\}. From these three we always report the
experiment with the highest validation log-likelihood. In the majority of our
experiments a learning rate of 0.001 gave the best results, even across
different layer models (SBN, AR-SBN and NADE). If not noted otherwise, we use
$K=5$ samples during training and $K=100,000$ samples to estimate the final
log-likelihood on the test set\footnote{We refer to the lower bound estimates which
can be arbitrarily tightened by increasing the number of test samples as
{\em LL estimates} to distiguish them from the {\em variational LL
lower bounds} (see section \ref{sec:pxest}).}.
To disentangle the influence of the different q updating methods we setup $p$
and $q$ networks consisting of three hidden SBN layers with 10, 200
and 200 units (SBN/SBN 10-200-200). After convergence, the model trained
updating q during the sleep phase only reached a final estimated
log-likelihood of $-93.4$, the model trained with a q-update during the wake
phase reached $-92.8$, and the model trained with both wake and sleep phase
update reached $-91.9$. As a control we trained a model that does not update
$q$ at all. This model reached $-171.4$. We confirmed that combining wake
and sleep phase q-updates generally gives the best results by repeating this
experiment with various other architectures. For the remainder of this paper we
therefore train all models with combined wake and sleep phase q-updates.

Next we investigate the influence of the number of samples used during
training. The results are visualized in Fig. \ref{fig:mnist1} A. Although the
results depend on the layer-distributions and on the depth and width of the
architectures, we generally observe that the final estimated log-likelihood
does not improve significantly when using more than 5 samples during training 
for NADE models, and using more than 25 samples for models with SBN layers.
We can quantify the bias and the variance of the gradient estimator
(\ref{eq:grad_p}) using bootstrapping. While training a SBN/SBN 10-200-200 model
with $K=100$ training samples, we use $K=5,000$ samples to get a high quality
estimate of the gradient for a small but fixed set of 25 datapoints (the size
of one mini-batch). By repeatedly resampling smaller sets of $\{1, 2, 5, \cdots,
5000\}$ samples with replacement and by computing the gradient based on these, we get a measure for
the bias and the variance of the small sample estimates relative the high
quality estimate. These results are visualized in Fig. \ref{fig:mnist1} B.
In Fig. \ref{fig:mnist2} A we finally investigate the quality of the 
log-likelihood estimator (eqn. \ref{eq:pxest}) when applied to the
MNIST test set.

Table \ref{tab:mnist} summarizes how different architectures compare to each
other and how RWS compares to related methods for training directed models. 
We essentially observe that RWS trained models consistently improve over 
classical wake-sleep, especially for deep architectures. We 
furthermore observe that using autoregressive layers (AR-SBN or NADE) for the inference network 
improves the results even when the generative model is composed of factorial SBN layers.
Finally, we see that the best performing models with autoregressive layers in $p$
are always shallow with only a single hidden layer.
In Table \ref{tab:results} (left) we compare some of our best models to
the state-of-the-art results published on MNIST. The deep SBN/SBN
10-100-200-300-400 model was trained for 1000 epochs with $K=5$ training
samples and a learning rate of $0.001$. For fine-tuning we run additional 500 epochs
with a learning rate decay of $1.005$ and 100 training samples.
For comparison we also train the best performing model from the
DARN paper \citep{KarolAndriyDaan14} with RWS, i.e., a single layer AR-SBN with
500 latent variables and a {\em deterministic} layer of hidden variables
between the observed and the latents. We essentially obtain the same final 
testset log-likelihood. For this shallow network we thus do not observe 
any improvement from using RWS.

%training samples
%and additional our model with the largest log-likelihood reaches
%$85.32$ and is a shallow model composed of (conditional) NADEs with 250
%hidden units. This model was trained using 50 epochs with a learning rate of $0.003$
%and $K$=5 samples and another $50$ epochs with a learning rate of $0.001$ and
%$K$=25 training samples. 

\vspace{-0.2cm}
\subsection{CalTech 101 Silhouettes}
\vspace{-0.2cm}

\begin{figure}[t]
  \centering
  \includegraphics[width=\linewidth]{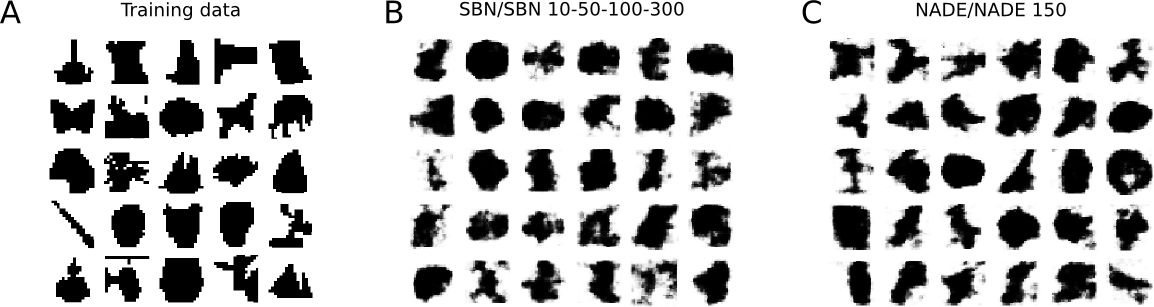}
  \vspace{-0.5cm}
  \caption{
    CalTech 101 Silhouettes:
    {\bf A} Random selection of training data points.
    {\bf B} Random samples from the SBN/SBN 10-50-100-300 generative network.
    {\bf C} Random Samples from the NADE-150 generative network.
    (We show the probabilities from which each pixel is sampled)
  }
  \label{fig:caltech}
\end{figure}

We applied reweighted wake-sleep to the $28 \times 28$ pixel CalTech 101
Silhouettes dataset. This dataset consists of 4,100 examples in the training
set, 2,264 examples in the validation set and 2,307 examples in the test set.  We
trained various architectures on this dataset using the same hyperparameter as
for the MNIST experiments. Table \ref{tab:results} (right) summarizes our results.
Note that our best SBN/SBN model is a relatively deep network with 4 
hidden layers (300-100-50-10) and reaches a estimated LL of -116.9 on the test set.
Our best network, a shallow NADE/NADE-150 network reaches -104.3 and
improves over the previous state of the art ($-107.8$, a RBM with 4000 hidden 
units by~\cite{Cho2013enhanced}).

\vspace{-0.2cm}
\section{Conclusions}
\vspace{-0.2cm}

We introduced a novel training procedure for deep generative models consisting
of multiple layers of binary latent variables. It generalizes and improves over
the wake-sleep algorithm providing a lower bias and lower variance estimator of
the log-likelihood gradient at the price of more samples from the inference
network. During training the weighted samples from the inference network decouple 
the layers such that the learning gradients only propagate within the individual layers.
Our experiments demonstrate that a small number of $\approx 5$ samples 
is typically sufficient to jointly train relatively deep architectures of at least 5
hidden layers without layerwise pretraining and without carefully tuning
learning rates.  The resulting models produce reasonable samples (by visual
inspection) and they approach state-of-the-art performance in terms of
log-likelihood on several discrete datasets. 

We found that even in the cases when the generative networks contain SBN layers
only, better results can be obtained with inference networks composed of more
powerful, autoregressive layers. This however comes at the price of reduced
computational efficiency on e.g. GPUs as the individual variables $h_i \sim q(h|\xVec)$ have to
be sampled in sequence (even though the theoretical complexity is not
significantly worse compared to SBN layers).

We furthermore found that models with autoregressive layers in the generative
network $p$ typically produce very good results. But the best ones were always
shallow with only a single hidden layer. At this point it is unclear if this is
due to optimization problems. 

{
%\small
{\bf Acknowledgments} \;\;
%\vspace{-0.3cm}
%\subsubsection*{Acknowledgments} 
%\vspace{-0.3cm}
We would like to thank Laurent Dinh, Vincent Dumoulin and Li Yao for helpful discussions and the 
developers of Theano~\citep{bergstra+al:2010-scipy,Bastien-Theano-2012} for their powerful software.
We furthermore acknowledge CIFAR and Canada Research Chairs for funding and Compute Canada, and Calcul Qu\'ebec
for providing computational resources.
}

\bibliography{strings,strings-shorter,ml,aigaion,localref}

\begin{thebibliography}{}

\bibitem[Bastien {\em et~al.}(2012)Bastien, Lamblin, Pascanu, Bergstra,
  Goodfellow, Bergeron, Bouchard, and Bengio]{Bastien-Theano-2012}
Bastien, F., Lamblin, P., Pascanu, R., Bergstra, J., Goodfellow, I.~J.,
  Bergeron, A., Bouchard, N., and Bengio, Y. (2012).
\newblock Theano: new features and speed improvements.
\newblock Deep Learning and Unsupervised Feature Learning NIPS 2012 Workshop.

\bibitem[Bengio(2009)Bengio]{Bengio-2009-book}
Bengio, Y. (2009).
\newblock {\em Learning deep architectures for {AI}\/}.
\newblock Now Publishers.

\bibitem[Bengio and Bengio(2000)Bengio and Bengio]{Bengio+Bengio-NIPS99}
Bengio, Y. and Bengio, S. (2000).
\newblock Modeling high-dimensional discrete data with multi-layer neural
  networks.
\newblock In {\em NIPS'99\/}, pages 400--406. MIT Press.

\bibitem[Bengio {\em et~al.}(2013)Bengio, Mesnil, Dauphin, and
  Rifai]{Bengio-et-al-ICML2013}
Bengio, Y., Mesnil, G., Dauphin, Y., and Rifai, S. (2013).
\newblock Better mixing via deep representations.
\newblock In {\em Proceedings of the 30th International Conference on Machine
  Learning (ICML'13)\/}. ACM.

\bibitem[Bergstra {\em et~al.}(2010)Bergstra, Breuleux, Bastien, Lamblin,
  Pascanu, Desjardins, Turian, Warde-Farley, and
  Bengio]{bergstra+al:2010-scipy}
Bergstra, J., Breuleux, O., Bastien, F., Lamblin, P., Pascanu, R., Desjardins,
  G., Turian, J., Warde-Farley, D., and Bengio, Y. (2010).
\newblock Theano: a {CPU} and {GPU} math expression compiler.
\newblock In {\em Proceedings of the Python for Scientific Computing Conference
  ({SciPy})\/}.
\newblock Oral Presentation.

\bibitem[Boulanger-Lewandowski {\em et~al.}(2012)Boulanger-Lewandowski, Bengio,
  and Vincent]{Boulanger+al-ICML2012-small}
Boulanger-Lewandowski, N., Bengio, Y., and Vincent, P. (2012).
\newblock Modeling temporal dependencies in high-dimensional sequences:
  Application to polyphonic music generation and transcription.
\newblock In {\em ICML'2012\/}.

\bibitem[Cho {\em et~al.}(2013)Cho, Raiko, and Ilin]{Cho2013enhanced}
Cho, K., Raiko, T., and Ilin, A. (2013).
\newblock Enhanced gradient for training restricted boltzmann machines.
\newblock {\em Neural computation\/}, {\bf 25}(3), 805--831.

\bibitem[Dayan and Hinton(1996)Dayan and Hinton]{Dayan1996varieties}
Dayan, P. and Hinton, G.~E. (1996).
\newblock Varieties of helmholtz machine.
\newblock {\em Neural Networks\/}, {\bf 9}(8), 1385--1403.

\bibitem[Dayan {\em et~al.}(1995)Dayan, Hinton, Neal, and
  Zemel]{Dayan-et-al-1995}
Dayan, P., Hinton, G.~E., Neal, R.~M., and Zemel, R.~S. (1995).
\newblock The {H}elmholtz machine.
\newblock {\em Neural computation\/}, {\bf 7}(5), 889--904.

\bibitem[Frey(1998)Frey]{Frey98}
Frey, B.~J. (1998).
\newblock {\em Graphical models for machine learning and digital
  communication\/}.
\newblock {MIT} Press.

\bibitem[Gregor {\em et~al.}(2014)Gregor, Danihelka, Mnih, Blundell, and
  Wierstra]{KarolAndriyDaan14}
Gregor, K., Danihelka, I., Mnih, A., Blundell, C., and Wierstra, D. (2014).
\newblock Deep autoregressive networks.
\newblock In {\em Proceedings of the 31st International Conference on Machine
  Learning\/}.

\bibitem[Hinton {\em et~al.}(1995)Hinton, Dayan, Frey, and Neal]{Hinton95}
Hinton, G.~E., Dayan, P., Frey, B.~J., and Neal, R.~M. (1995).
\newblock The wake-sleep algorithm for unsupervised neural networks.
\newblock {\em Science\/}, {\bf 268}, 1558--1161.

\bibitem[Hinton {\em et~al.}(2006)Hinton, Osindero, and Teh]{Hinton06}
Hinton, G.~E., Osindero, S., and Teh, Y. (2006).
\newblock A fast learning algorithm for deep belief nets.
\newblock {\em Neural Computation\/}, {\bf 18}, 1527--1554.

\bibitem[Kingma and Welling(2014)Kingma and Welling]{Kingma+Welling-ICLR2014}
Kingma, D.~P. and Welling, M. (2014).
\newblock Auto-encoding variational bayes.
\newblock In {\em Proceedings of the International Conference on Learning
  Representations (ICLR)\/}.

\bibitem[Larochelle(2011)Larochelle]{LarochelleBinarizedMNIST}
Larochelle, H. (2011).
\newblock Binarized mnist dataset.
\newblock \small
  \url{http://www.cs.toronto.edu/~larocheh/public/datasets/binarized_mnist/binarized_mnist_[train|valid|test].amat}.

\bibitem[Larochelle and Murray(2011)Larochelle and
  Murray]{Larochelle+Murray-2011}
Larochelle, H. and Murray, I. (2011).
\newblock The {N}eural {A}utoregressive {D}istribution {E}stimator.
\newblock In {\em Proceedings of the Fourteenth International Conference on
  Artificial Intelligence and Statistics (AISTATS'2011)\/}, volume 15 of JMLR:
  W\&CP.

\bibitem[Mnih and Gregor(2014)Mnih and Gregor]{Andriy-Karol-2014}
Mnih, A. and Gregor, K. (2014).
\newblock Neural variational inference and learning in belief networks.
\newblock In {\em Proceedings of the 31st International Conference on Machine
  Learning (ICML 2014)\/}.
\newblock to appear.

\bibitem[Murray and Larochelle(2014)Murray and Larochelle]{Uria+al-ICML2014}
Murray, B. U.~I. and Larochelle, H. (2014).
\newblock A deep and tractable density estimator.
\newblock In {\em ICML'2014\/}.

\bibitem[Murray and Salakhutdinov(2009)Murray and Salakhutdinov]{MurraySal09}
Murray, I. and Salakhutdinov, R. (2009).
\newblock Evaluating probabilities under high-dimensional latent variable
  models.
\newblock In {\em NIPS'08\/}, volume~21, pages 1137--1144.

\bibitem[Rezende {\em et~al.}(2014)Rezende, Mohamed, and
  Wierstra]{Rezende-et-al-ICML2014}
Rezende, D.~J., Mohamed, S., and Wierstra, D. (2014).
\newblock Stochastic backpropagation and approximate inference in deep
  generative models.
\newblock In {\em ICML'2014\/}.

\bibitem[Salakhutdinov and Murray(2008)Salakhutdinov and Murray]{SalMurray08}
Salakhutdinov, R. and Murray, I. (2008).
\newblock On the quantitative analysis of deep belief networks.
\newblock In {\em Proceedings of the International Conference on Machine
  Learning\/}, volume~25.

\bibitem[Saul {\em et~al.}(1996)Saul, Jaakkola, and Jordan]{Saul+96}
Saul, L.~K., Jaakkola, T., and Jordan, M.~I. (1996).
\newblock Mean field theory for sigmoid belief networks.
\newblock {\em Journal of Artificial Intelligence Research\/}, {\bf 4}, 61--76.

\bibitem[Tang and Salakhutdinov(2013)Tang and
  Salakhutdinov]{Tang+Salakhutdinov-NIPS2013}
Tang, Y. and Salakhutdinov, R. (2013).
\newblock Learning stochastic feedforward neural networks.
\newblock In {\em NIPS'2013\/}.

\end{thebibliography}
\bibliographystyle{natbib}
%\bibliographystyle{unsrt}

%%%%%%%%%%%%%%%%%%%%%%%%%%%%%%%%%%%%%%%%%%%%%%%%%%%%%%%%%%%%%%%%%%%%%%%%%%%%%%%%
\newpage
\section{Supplement}

\subsection{Gradients for $p(\xVec, \hVec)$}
\begin{align}
  \label{eq:grad_p_detailed}
  \frac{\partial}{\partial \theta} \LL_p(\theta, \xVec \sim \DD) &=
       \frac{\partial}{\partial \theta} \log p_{\theta}(\xVec) 
%     = \frac{1}{p(\xVec)} \frac{\partial}{\partial \theta} p(\xVec) \\
     = \frac{1}{p(\xVec)} \frac{\partial}{\partial \theta} \sum_{\hVec} p(\xVec, \hVec) \nonumber \\
    &= \frac{1}{p(\xVec)} \sum_{\hVec} p(\xVec, \hVec) 
            \frac{\partial}{\partial \theta} \log  p(\xVec, \hVec) \nonumber \\
    &= \frac{1}{p(\xVec)} \sum_{\hVec} \qrob{\hVec}{\xVec} \frac{p(\xVec, \hVec)}{\qrob{\hVec}{\xVec}}
            \frac{\partial}{\partial \theta} \log  p(\xVec, \hVec) \nonumber \\
    &= \frac{1}{p(\xVec)} \E{\hVec \sim \qrob{\hVec}{\xVec}}{\frac{p(\xVec, \hVec)}{\qrob{\hVec}{\xVec}}
            \frac{\partial}{\partial \theta} \log p(\xVec, \hVec)} \nonumber \\
    &\simeq \frac{1}{\sum_k \omega_k} \sum_{k=1}^K \omega_k
            \frac{\partial}{\partial \theta} \log p(\xVec, \hVec^{(k)}) \\
\text{with}\;&\omega_k = \frac{p(\xVec, \hVec^{(k)})}{\qrob{\hVec^{(k)}}{\xVec}} \;
\text{and}\; \hVec^{(k)} \sim \qrob{\hVec}{\xVec} \nonumber
\end{align}

\subsection{Gradients for the wake phase q update}
\begin{align}
  \label{eq:grad_q_detailed}
  \frac{\partial}{\partial \phi} \LL_q(\phi, \xVec \sim \DD) 
    &= \frac{\partial}{\partial \phi} \sum_{\hVec} 
        p(\xVec, \hVec) {\log q_{\phi}(\hVec | \xVec)}  \nonumber \\
    &= \frac{1}{p(\xVec)} \E{\hVec \sim \qrob{\hVec}{\xVec}}{\frac{p(\xVec, \hVec)}{\qrob{\hVec}{\xVec}}
            \frac{\partial}{\partial \phi} \log q_{\phi}(\hVec | \xVec)}  \nonumber \\
    &\simeq \frac{1}{\sum_k \omega_k} \sum_{k=1}^K \omega_k
            \frac{\partial}{\partial \phi} \log q_{\phi}(\hVec^{(k)} | \xVec)
%\text{with}\;&\omega_k = \frac{p(\xVec, \hVec^{(k)})}{\qrob{\hVec^{(k)}}{\xVec}} \;
%\text{and}\; \hVec^{(k)} \sim \qrob{\hVec}{\xVec} \nonumber
%    &= \cdots \\
%    &= \E{\xVec \sim \DD}{ \E{\hVec \sim \qrob{\hVec}{\xVec)}} 
%     { \omega \; \frac{\partial}{\partial \phi} \log \qrob{\hVec}{\xVec}} }
%     \; \text{with the importance weights } \omega = \frac{\prob{\hVec}{\xVec}}{\qrob{\hVec}{\xVec}}
\end{align}
Note that we arrive at the same gradients when we set out to minimize
the $KL(p(\cdot|\xVec) \parallel q(\cdot|\xVec)$ for a given datapoint $\xVec$:
\begin{align}
  \frac{\partial}{\partial \phi} KL (p_{\theta}(\hVec | \xVec) \parallel q_{\phi}(\hVec | \xVec)) 
  &= \frac{\partial}{\partial \phi} 
     \sum_{\hVec} p_{\theta}(\hVec|x) \log \frac{p_{\theta}(\hVec|\xVec)}{q_{\phi}(\hVec|\xVec)} \nonumber \\
  &= - \sum_{\hVec} p_{\theta}(\hVec|\xVec) \frac{\partial}{\partial \phi} \log q_{\phi}(\hVec|\xVec) \nonumber \\
  &\simeq - \frac{1}{\sum_k \omega_k} \sum_{k=1}^K \omega_k
     \frac{\partial}{\partial \phi} \log q_{\phi}(\hVec^{(k)} | \xVec)
\end{align} 
\\

\newpage

\subsection{Additional experimental results}

\subsubsection{Learning curves for MNIST experiments}
\begin{figure}[h!]
  \centering
  \includegraphics[width=0.7\linewidth]{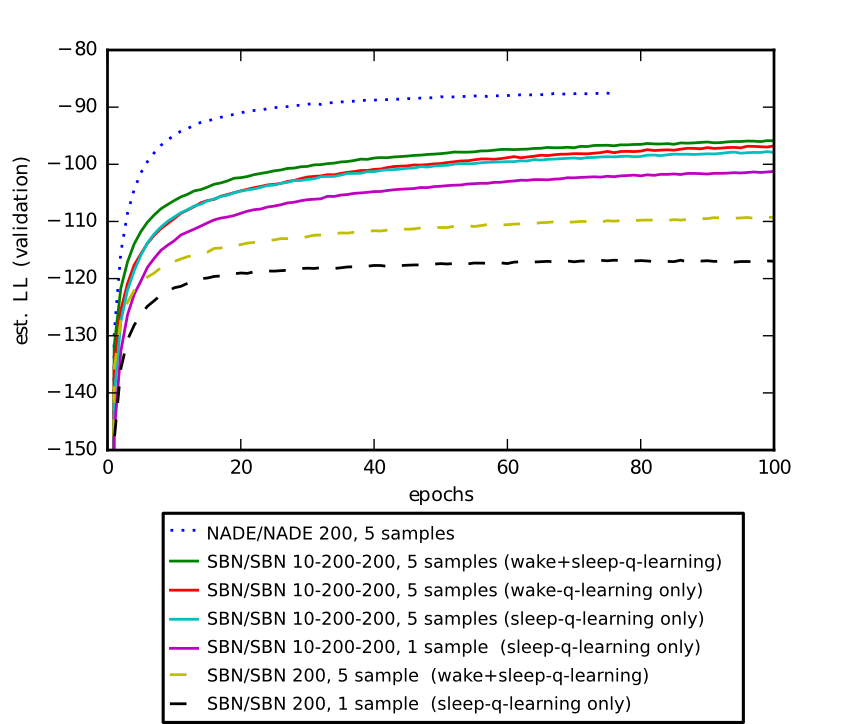}
  \vspace{-0.5cm}
  \caption{
    Learning curves for various MNIST experiments.
  }
\end{figure}

\subsubsection{Bootstrapping based $\log(p(x))$ bias/variance analysis}

Here we show the bias/variance analysis from Fig. 1 B (main paper) 
applied to the estimated $\log(p(x))$ w.r.t. the number of test samples. \\

\begin{figure}[h!]
  \centering
  \includegraphics[width=0.7\linewidth]{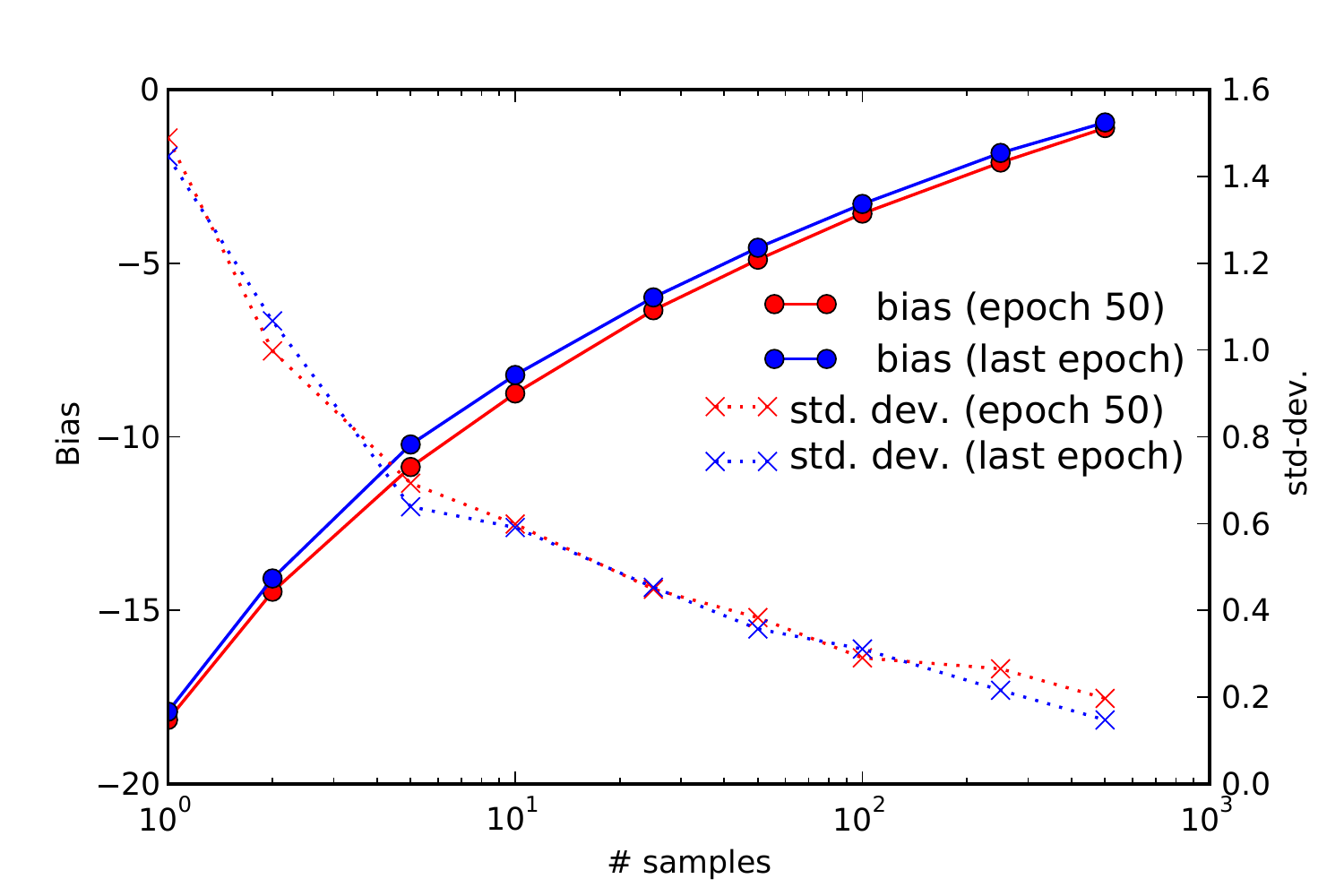}
  \vspace{-0.5cm}
  \caption{
    Bias and standard deviation of the low-sample estimated $\log(p(x))$
    (bootstrapping with K=5,000 primary samples from a SBN/SBN 10-200-200 
    network trained on MNIST).
  }
\end{figure}

\newpage

\subsubsection{UCI binary datasets}

We performed a series of experiments on 8 different binary datasets from the
UCI database:

For each dataset we screened a limited hyperparameter space: The learning rate
was set to a value in ${0.001, 0.003, 0.01}$. For SBNs we use $K$=10
training samples and we tried the following architectures: Two hidden layers
with 10-50, 10-75, 10-100, 10-150 or 10-200 hidden units and three hidden
layers with 5-20-100, 10-50-100, 10-50-150, 10-50-200 or 10-100-300 hidden
units. We trained NADE/NADE models with $K$=5 training samples and one hidden
layer with 30, 50, 75, 100 or 200 units in it. 
\begin{table}[h!]
\centering
\scriptsize
%\tiny
\begin{tabular}{|l|c|c|c|c|c|c|c|c|}
\hline
Model         & \tiny \bf ADULT & \tiny \bf CONNECT4 & \tiny \bf DNA & \tiny \bf MUSHROOMS & \tiny \bf NIPS-0-12 & \tiny \bf OCR-LETTERS & \tiny \bf RCV1 & \tiny \bf WEB \\
\hline
\hline
{\bf FVSBN}                & 13.17    & 12.39     & 83.64       & 10.27       & 276.88 & 39.30 & 49.84 & 29.35 \\
{\bf NADE$^{*}$}           & 13.19    & 11.99     & 84.81       &  9.81       & 273.08 & 27.22 & 46.66 & 28.39 \\
{\bf EoNADE$^{+}$}         & 13.19    & 12.58     & 82.31       &  9.68       & 272.38 & 27.31 & 46.12 & {\bf 27.87} \\
{\bf DARN$^{3}$}           & 13.19    & 11.91     & {\bf 81.04} &  {\bf 9.55} & 274.68 & 28.17 & 46.10 & 28.83 \\
\hline
{\bf RWS - SBN}            & 13.65    & 12.68     &  90.63      &  9.90       & 272.54    & 29.99      & 46.16     & 28.18 \\
{\tiny hidden units}       & 5-20-100 & 10-50-150 & 10-150      & 10-50-150   & 10-50-150 & 10-100-300 & 10-50-200 & 10-50-300 \\
%{\bf RWS - DARN}           &       &       &       &       &        &       &       & \\
\hline
{\bf RWS - NADE}           & {\bf 13.16} & {\bf 11.68} & 84.26 &  9.71 & {\bf 271.11} & {\bf 26.43} & {\bf 46.09} & 27.92 \\
{\tiny hidden units}       & 30          &  50         &  100  &   50  &  75    & 100   &    &  \\
\hline
\end{tabular}
\caption{
Results on various binary datasets from the UCI repository. The top two rows
quote the baseline results from Larochelle \& Murray (2011); the third row
shows the baseline results taken from Uria, Murray, Larochelle (2014).
(NADE$^{*}$: 500 hidden units; EoNADE$^{+}$: 1hl, 16 ord)
}
\label{tab:other}
\end{table}

\end{document}